\documentclass[sigconf, screen]{acmart}

\usepackage{multirow} 
\usepackage{graphicx}
\usepackage{float}

\AtBeginDocument{%
  }
    
\renewcommand\footnotetextcopyrightpermission[1]{}
\acmSubmissionID{4684}

\begin{document}

\title{AttentionGS: Towards Initialization-Free 3D Gaussian Splatting via Structural Attention}

\author{Ziao Liu}
\email{2019302141299@whu.edu.cn}
\affiliation{%
  \institution{Wuhan University}
  \city{Wuhan}
  \state{Hubei}
  \country{China}
}

\author{Zhenjia Li}
\email{lizhenjia003@ke.com}
\affiliation{%
  \institution{BEKE.inc}
  \city{Beijing}
  \state{Beijing}
  \country{China}
}

\author{Yifeng Shi}
\email{shiyifeng003@ke.com}
\affiliation{%
  \institution{BEKE.inc}
  \city{Beijing}
  \state{Beijing}
  \country{China}
}

\author{Xiangang Li}
\email{lixiangang002@ke.com}
\affiliation{%
  \institution{BEKE.inc}
  \city{Beijing}
  \state{Beijing}
  \country{China}
}

\begin{abstract}
3D Gaussian Splatting (3DGS) is a powerful alternative to Neural Radiance Fields (NeRF), excelling in complex scene reconstruction and efficient rendering. However, it relies on high-quality point clouds from Structure-from-Motion (SfM), limiting its applicability. SfM also fails in texture-deficient or constrained-view scenarios, causing severe degradation in 3DGS reconstruction. To address this limitation, we propose AttentionGS, a novel framework that eliminates the dependency on high-quality initial point clouds by leveraging structural attention for direct 3D reconstruction from randomly initialization. In the early training stage, we introduce geometric attention to rapidly recover the global scene structure. As training progresses, we incorporate texture attention to refine fine-grained details and enhance rendering quality. Furthermore, we employ opacity-weighted gradients to guide Gaussian densification, leading to improved surface reconstruction. Extensive experiments on multiple benchmark datasets demonstrate that AttentionGS significantly outperforms state-of-the-art methods, particularly in scenarios where point cloud initialization is unreliable. Our approach paves the way for more robust and flexible 3D Gaussian Splatting in real-world applications.
\end{abstract}



\begin{teaserfigure}
  \centering
  \includegraphics[width=\linewidth]{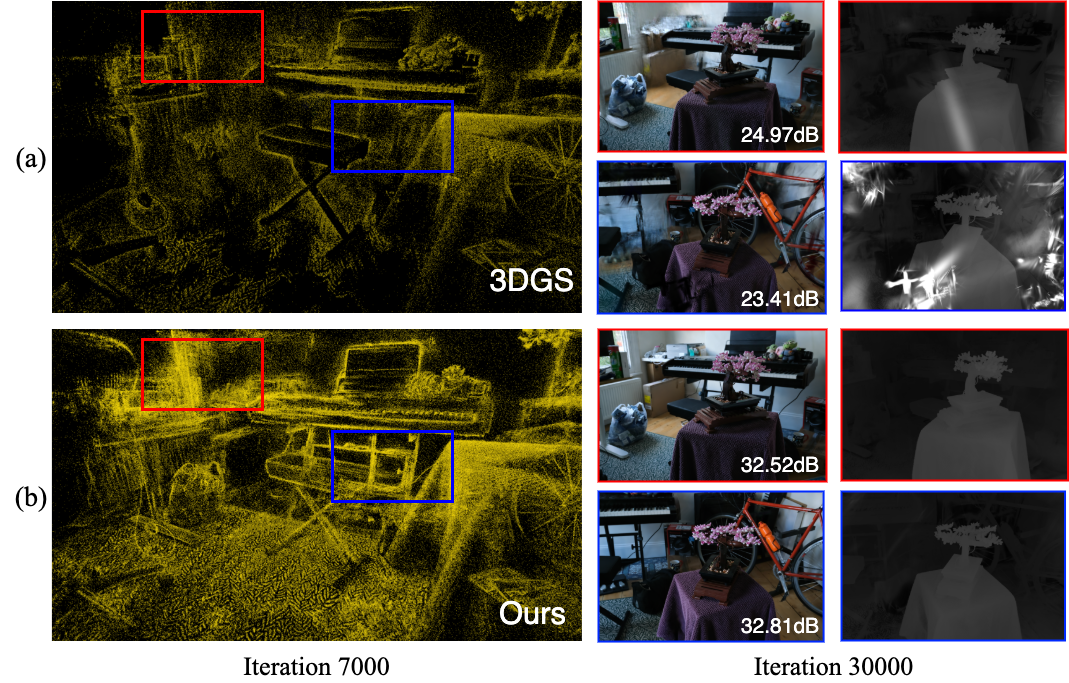}
  \caption{The comparison between 3DGS and our method in terms of structural recovery is illustrated, with both approaches utilizing random point cloud initialization. The results after 7000 iterations are presented. We use blue dots to represent Gaussian centers, demonstrating whether the Gaussian points are correctly distributed.}
  \label{fig:abstract}
\end{teaserfigure}

\maketitle

\section{Introduction}

Novel view synthesis stands as a pivotal challenge in the domain of 3D vision, focusing on the task of inferring continuous geometric and appearance representations of a scene from sparse 2D observations. Although neural radiance fields\cite{nerf,nerf01,nerf02,nerf03,nerf04} (NeRF) have achieved remarkable rendering quality through implicit neural representations, their reliance on multilayer perceptron architectures inherently limits computational efficiency, posing significant challenges for real-time interaction. In contrast, 3D Gaussian Splatting\cite{3dgs,mipgs,pixelgs} (3DGS) introduces an innovative approach by combining explicit Gaussian primitive parameterization with a differentiable rasterization pipeline. This method achieves real-time rendering capabilities while maintaining high-quality reconstruction, thereby paving the way for advanced scene modeling.

3D Gaussian Splatting (3DGS) typically requires initialization with high-quality point clouds, which are commonly obtained using laser scanning. However, such methods incur substantial financial and operational costs, limiting their practicality for large-scale deployment. A more widely adopted alternative is to generate point clouds from images using Structure-from-Motion (SfM)-based approaches, such as COLMAP. While these methods still introduce considerable computational overhead,which can even rival the computational cost of training 3DGS itself. Furthermore, in scenes with low-texture regions or complex geometries, SfM often fails to reconstruct satisfactory point clouds, posing significant challenges to the overall 3DGS reconstruction process.

In response to the limitations posed by the dependence on high-quality initial point clouds, we introduce AttentionGS, a novel framework designed to perform direct 3D reconstruction from random initialization by leveraging structural attention mechanisms. As shown in Fig\ref{fig:abstract}(b), our method achieves better structural reconstruction in the early stages of training, leading to further improvements in rendering quality. AttentionGS first integrates a \textbf{Geometric Attention Module} to ensure robust structural fidelity. This module accelerates structure learning near object boundaries and enhances the recovery of geometric details from random initialization. An \textbf{Opacity-Weighted Gradient Strategy} is then introduced to further capture and represent surface details accurately, which improves the reconstruction of surface structures by refining the densification mechanism. Finally, to mitigate the delayed reconstruction of appearance attributes resulting from the emphasis on structural learning, we introduce an \textbf{Appearance Attention Module}. By leveraging channel-specific weights in the RGB space, this module aligns more effectively with the Gaussian color representation, enabling rapid appearance recovery once the scene structure has been accurately reconstructed.To summarize, the main contributions of this work are as follows:
\begin{itemize}
\item We propose a Geometric Attention Module and an Opacity-Weighted Gradient Strategy to resolve the contradiction between structural recovery and the optimization of photometric attributes in 3DGS, enhancing overall reconstruction and rendering performance.

\item We introduce an Appearance Attention Module to enable rapid recovery of appearance attributes after the structure has been accurately reconstructed.

\item Our method achieves state-of-the-art performance on the challenging Mip-NeRF 360 and LLFF datasets under conditions of random and noisy point cloud initialization, demonstrating its ability to eliminate dependency on high-quality point clouds.
\end{itemize}

\section{Related Work}
\subsection{Structure from Motion}
Structure-from-Motion (SfM) techniques \cite{sfm0,sfm1,sfm2} have become one of the most widely used algorithms for reconstructing 3D scenes due to their ability to generate accurately aligned point clouds and precise camera poses. Typical SfM methods output the pose for each input image and a sparse point cloud containing rough color and position information for each point. Despite their effectiveness, the incremental nature of SfM and the computational intensity of the bundle adjustment process significantly increase its time complexity, often reaching $O(n^4)$ with respect to the number of cameras involved. 

To address these limitations, learning-based methods such as DUSt3R \cite{wang2024dust3r,zhang2024monst3r}have been developed to enhance the capabilities of 3D reconstruction from images. DUSt3R leverages deep learning techniques to perform Dense Unconstrained Stereo 3D Reconstruction, which allows it to generate accurate 3D models from un-calibrated and un-posed cameras.  This is achieved through a neural network that learns to map 2D image patterns to 3D shapes in a unified model, eliminating the need for explicit triangulation.However, despite these advancements, DUSt3R still faces challenges when it comes to processing a large number of images simultaneously. The computational demands and memory requirements can become significant, limiting its scalability for very large datasets. Additionally, there is a degree of scale distortion that can occur, which affects the absolute accuracy of the reconstructed 3D models.
\subsection{Novel view synthesis}
Neural Radiance Fields (NeRF) and its derivatives have made significant advancements in novel view synthesis (NVS), setting new standards for rendering high-quality images from novel viewpoints. NeRF methods achieve impressive visual fidelity through volumetric scene representations \cite{nerf}. Building on this foundation, some works focus on optimizing camera positions and poses to enhance viewpoint synthesis accuracy \cite{nerf01,nerf02,nerf03,nerf04}. Concurrently, other studies aim to improve rendering quality by refining neural network architectures to enhance image detail and realism \cite{nerf11,nerf12,nerf13}. Additionally, to meet the demands of real-time applications, certain efforts concentrate on speed optimization, employing algorithmic and hardware acceleration to reduce computational resource consumption \cite{nerf21}.

Despite the significant visual advancements of NeRF methods, they often require extensive computational resources and dense sampling, limiting their applicability in real-time scenarios. In contrast, 3DGS \cite{3dgs} offers a more efficient solution by representing scenes with Gaussian functions. This method excels in rendering speed and detail preservation, making it a compelling alternative to traditional NeRF approaches. Recent advancements in 3DGS have further improved its capabilities. Some works have focused on anti-aliasing techniques to enhance visual quality \cite{pixelgs,mvsplat,mipgs}, while others have developed methods to reduce storage requirements, making 3DGS more practical for large-scale applications \cite{xie2024gaussiancity}. These developments highlight the ongoing efforts to refine 3DGS, addressing its limitations and expanding its applicability. However, 3DGS's heavy reliance on the quality of the initial point cloud poses a significant limitation in practical applications. This dependency motivates the development of our method, which aims to mitigate the reliance on high-quality initial point clouds while maintaining the rendering and detail advantages of 3DGS.

\begin{figure*}[h]
  \centering
  \includegraphics[width=\linewidth]{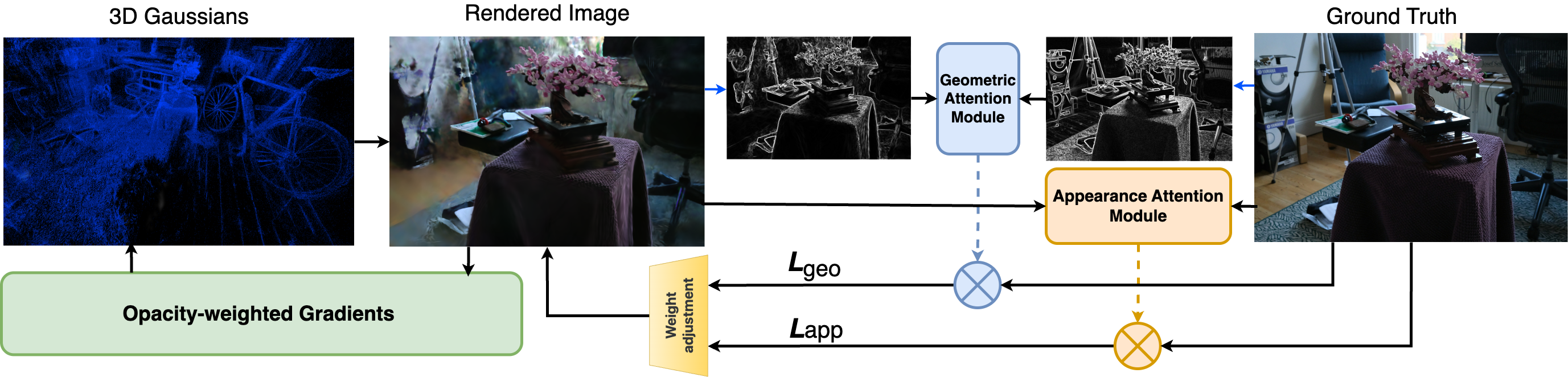}
  \caption{Overview of AttentionGS.}
  \label{fig:overview}
\end{figure*}

\section{Methodology}

In response to the limitations posed by the dependency on high-quality initial point clouds, we introduce AttentionGS, a novel framework designed to perform direct 3D reconstruction from random initialization. This approach leverages structural attention mechanisms, thereby obviating the necessity for high-quality initial point clouds. Our methodology is structured around three pivotal components: the geometric attention module, the opacity-weighted gradient strategy, and the appearance attention module.

\subsection{Preliminary}

3DGS is a sophisticated technique employed to efficiently represent and render 3D scenes through the utilization of Gaussian functions. Each Gaussian is defined by its mean position, covariance, and radiance properties, collectively approximating the continuous radiance field of a scene. This method is particularly effective in managing complex geometries and lighting conditions inherent in 3D reconstruction tasks.

Mathematically, a Gaussian function $G$ in 3D space is expressed as:

\begin{equation}
G(\mathbf{x}) = \exp\left(-\frac{1}{2} (\mathbf{x} - \boldsymbol{\mu})^\top \boldsymbol{\Sigma}^{-1} (\mathbf{x} - \boldsymbol{\mu})\right)
\end{equation}

where $\mathbf{x}$ represents a point in 3D space, $\boldsymbol{\mu}$ is the mean vector indicating the Gaussian's center, and $\boldsymbol{\Sigma}$ is the covariance matrix that defines the shape and orientation of the Gaussian.

A critical aspect of the 3DGS framework is the Gaussian densification strategy, which facilitates adaptive refinement of the scene representation. Within this framework, the decision to split a Gaussian is contingent upon the average magnitude of the gradient of the normalized device coordinates (NDC) for the viewpoints in which the Gaussian is involved. Specifically, for a Gaussian $i$ under viewpoint $k$, the NDC coordinates are denoted as $\boldsymbol{\mu}_{ndc}^{i,k}$, and the associated loss under viewpoint $k$ is represented by $L_k$.

During the adaptive density control process, which is executed every predetermined number of iterations, Gaussian $i$ is evaluated across $M^i$ viewpoints. The threshold $\tau_{pos}$ is predefined to ascertain when a Gaussian should be split. The condition for densification is articulated as:

\begin{equation}
\frac{1}{M^i} \sum_{k=1}^{M^i} \left\| \nabla_{\boldsymbol{\mu}_{ndc}^{i,k}} L_k \right\| > \tau_{pos}
\end{equation}

where $\nabla_{\boldsymbol{\mu}_{ndc}^{i,k}} L_k$ denotes the gradient of the loss with respect to the NDC coordinates. If this condition is met, the Gaussian is bifurcated, allowing for a more nuanced representation of the scene.

\subsection{Geometric Attention Module}

Our observations indicate that Gaussian point clouds exhibit a propensity to adjust their color rather than their position when fitting images. This tendency can impede the accurate recovery of geometric structures, particularly when initialized with random point clouds. To mitigate this issue, we have developed the Geometric Attention Module.

Initially, we employ the Canny operator to extract edge details from RGB images. The Canny operator is defined as follows:

\begin{equation}
\text{Canny}(I) = \text{NonMaxSuppression}(\text{GradientMagnitude}(I))
\end{equation}

where $\text{GradientMagnitude}(I)$ signifies the gradient magnitude of image $I$, and $\text{NonMaxSuppression}$ is utilized to suppress non-maximum values. To further enhance the influence of detected edges, we apply an outward convolution operation on the output of the Canny operator. This convolution step is designed to expand the influence range of the edges, thereby allowing the edge information to impact a broader area of the image. The convolution operation can be mathematically represented as:

\begin{equation}
\text{EdgeEnhanced}(I) = \text{Convolve}(\text{Canny}(I), K)
\end{equation}

where $\text{Convolve}$ denotes the convolution operation and $K$ is the convolution kernel designed to expand the edge influence.

Subsequently, we compute and normalize the difference in edge details between the ground truth (GT) image and the RGB image. The weight calculation formula is given by:

\begin{equation}
w(x, y) = \frac{|\text{EdgeEnhanced}_{\text{GT}}(x, y) - \text{EdgeEnhanced}_{\text{RGB}}(x, y)|}{\max(|\text{EdgeEnhanced}_{\text{GT}} - \text{EdgeEnhanced}_{\text{RGB}}|)}
\end{equation}

where $\text{EdgeEnhanced}_{\text{GT}}$ and $\text{EdgeEnhanced}_{\text{RGB}}$ denote the enhanced edge detection results of the GT image and the RGB image, respectively, and $w(x, y)$ is the weight at position $(x, y)$.

The geometric weighted L1 loss is computed as:

\begin{equation}
L_{\text{geo}} = \sum_{x, y} w(x, y) \cdot |I_{\text{GT}}(x, y) - I_{\text{RGB}}(x, y)|
\end{equation}

This straightforward yet effective method aims to expedite the recovery of structures from random point clouds. Guided by the Geometric Attention Module, our method can achieve more efficient structural recovery with random point cloud initialization.

\subsection{Opacity-weighted Gradients Strategy}

To further augment the effectiveness of geometric recovery, we have refined the densification strategy of 3DGS. The original 3DGS calculates cumulative gradients using viewpoint-based weighting. While this approach effectively influences Gaussian distribution when the initial point cloud is of high quality, it may lead to similar densification tendencies between occluded and foreground Gaussians when initialized with random point clouds, thereby impeding accurate surface structure recovery.

To address this challenge, inspired by PixelGS, we incorporated the effect of opacity into the calculation of cumulative gradients. Specifically, we utilize the transmittance of each Gaussian at each pixel as a weighting factor to adjust the contribution of the gradient. The formula is articulated as follows:

\begin{equation}
\frac{\sum_{k=1}^{M^i} t_{i,k} \times \left\| \nabla_{\boldsymbol{\mu}_{ndc}^{i,k}} L_k \right\|}{\sum_{k=1}^{M^i} t_{i,k}} > \tau_{pos}
\end{equation}

where $t_{i,k}$ represents the transmittance of the Gaussian at pixel $k$, $\nabla_{\boldsymbol{\mu}_{ndc}^{i,k}} L_k$ is the gradient at viewpoint $k$, and $\tau_{pos}$ is the splitting threshold.

By incorporating transmittance as a weighting factor, we can more accurately distinguish between occluded and foreground Gaussians, allowing surface points to split more readily into correct positions, thereby significantly enhancing the precision of geometric structure recovery.

\subsection{Appearance Attention Module}

In the preceding sections, we successfully achieved rapid recovery of scene geometry using the Geometric Attention Module and Opacity-weighted Gradients Strategy. However, this strong emphasis on geometric structure may compromise the recovery of texture and color information. To address this issue, we propose the Appearance Attention Module to balance the recovery of geometric and appearance information.

The core of the Appearance Attention Module lies in adjusting weights based on the differences in RGB images. Specifically, we compute the absolute differences between the ground truth (GT) image and the RGB image and normalize them to serve as weights. Each color channel is calculated separately, as follows:

\begin{equation}
w_{\text{app}}(x, y, c) = \frac{|I_{\text{GT}}(x, y, c) - I_{\text{RGB}}(x, y, c)|}{\max(|I_{\text{GT}} - I_{\text{RGB}}|)}
\end{equation}

where $w_{\text{app}}(x, y, c)$ is the weight at position $(x, y)$ and channel $c$, and $I_{\text{GT}}$ and $I_{\text{RGB}}$ are the GT and RGB images, respectively.

The rationale for employing channel-specific weights in the RGB space is to further align with the Gaussian color representation. At this stage, the spherical harmonics used in our model have reached a relatively high order, which allows for a more nuanced representation of color variations. Calculating weights separately for each RGB channel, as opposed to using the mean pixel error across all channels as a uniform weight, offers superior convergence properties. This approach ensures that the subtle differences in color channels are captured more accurately, leading to a more precise reconstruction of the scene's appearance.

By focusing on channel-specific discrepancies, the Appearance Attention Module effectively enhances the model's ability to capture fine-grained color details. The separate treatment of RGB channels allows the model to adaptively adjust the influence of each channel based on its specific error, thereby improving the overall convergence rate and stability of the optimization process.

Based on these weights, we define the appearance-weighted L1 loss to better capture texture and color details:

\begin{equation}
L_{\text{app}} = \sum_{x, y, c} w_{\text{app}}(x, y, c) \cdot |I_{\text{GT}}(x, y, c) - I_{\text{RGB}}(x, y, c)|
\end{equation}

Inspired by the Sigmoid\cite{sigmoid}, we devised a dynamic adjustment function to progressively activate the Appearance Attention Module following geometric recovery. The formulation of this function is as follows:

\begin{equation}
f(i) = \frac{1}{1 + \exp\left(2 \times s \times \left(\frac{i}{N} - m\right)\right)}
\label{eq:m}
\end{equation}

where $i$ is the current iteration number, $s$ is the steepness parameter, $N$ is the total number of iterations, and $m$ is the decay node, determining the iteration point at which the Appearance Attention Module begins to significantly influence.

This dynamic adjustment ensures that the model initially focuses on recovering the geometric structure, which is crucial for establishing a solid foundation of the scene. As the iterations progress and the geometric structure becomes more stable, the influence of the Appearance Attention Module increases, allowing the model to refine the texture and color details without compromising the structural integrity.

Finally, the complete loss function combines the geometric and appearance attention modules, ensuring that while recovering geometric structures, the method also effectively maintains texture and color details:

\begin{equation}
L = L_1 + f(i) \cdot L_{\text{geo}} + (1 - f(i)) \cdot L_{\text{app}}
\end{equation}

By incorporating the Appearance Attention Module, our method can maintain texture and color recovery while preserving geometric accuracy. This comprehensive approach ensures a balanced recovery of both structural and appearance attributes, thereby enhancing the overall fidelity of the reconstructed 3D scenes. 

\section{Experiments}

\begin{table*}[h!]
    \scriptsize
    \caption{Quantitative results on the Mip-NeRF 360 dataset. For each metric, the best in {\textbf{bold}}.}
    \vspace{-2mm}
    \label{tab:main_result}
    \begin{center}
    \setlength\tabcolsep{2pt}
    \setlength{\belowcaptionskip}{0pt}
    \renewcommand\arraystretch{1.2}
    \scriptsize
    \vspace{-1.0em}
    \resizebox{0.95\linewidth}{!}{
        \begin{tabular}{c|l|ccc|ccc|ccc|ccc}
            \toprule
            && \multicolumn{3}{c}{Mip-NeRF 360(Avg.)} & \multicolumn{3}{c}{Bycicle} & \multicolumn{3}{c}{Bonsai} & \multicolumn{3}{c}{Stump}\\
            & & PSNR$\uparrow$ & SSIM$\uparrow$ & LPIPS$\downarrow$ & PSNR$\uparrow$ & SSIM$\uparrow$ & LPIPS$\downarrow$ & PSNR$\uparrow$ & SSIM$\uparrow$ & LPIPS$\downarrow$ & PSNR$\uparrow$ & SSIM$\uparrow$ & LPIPS$\downarrow$\\
            \midrule
            \multirow{4}{*}{Random}& 3DGS &24.89  &0.778  &0.167  & 22.06 & 0.562 & 0.366 &23.96  &0.841  &0.126 & 21.19 &0.609  &0.249\\
            & Pixel-GS &26.63  &0.829  &0.115  &23.67  &0.687 &0.223  &26.24  &0.865  &0.092 &22.21  &0.669  &0.189 \\
            & Mip-GS &24.61  & 0.789 &0.106  &22.48  &0.697  &\textbf{0.149}  &18.93  &0.628  &0.161  &21.12  &0.614  &0.228\\
            & Ours &\textbf{28.73}  &\textbf{0.866}  &\textbf{0.083}  &\textbf{25.13}  &\textbf{0.743}  &0.167  &\textbf{32.82}  &\textbf{0.953}  &\textbf{0.031} &\textbf{26.47}  &\textbf{0.761}  &\textbf{0.108}\\
            \specialrule{0.08em}{1pt}{1pt}
            \multirow{4}{*}{Noise}& 3DGS &27.09&0.819&0.127&23.87&0.697&0.191&29.83&0.923&0.054&24.38&0.672&0.212\\
            & Pixel-GS &26.48&0.828&0.114&24.51&0.739&0.161&25.05&0.841&0.121&24.94&0.721&0.145\\
            & Mip-GS &25.82&0.837&0.109&\textbf{25.65}&\textbf{0.802}&\textbf{0.093}&24.21&0.826&0.131&23.12&0.713&0.128\\
            & Ours &\textbf{28.79}&\textbf{0.867}&\textbf{0.078}&25.33&0.751&0.144&\textbf{32.51}&\textbf{0.948}&\textbf{0.034}&\textbf{26.77}&\textbf{0.762}&\textbf{0.105}\\
            \specialrule{0.08em}{1pt}{1pt}
            \multirow{2}{*}{Colmap}& 3DGS &28.86  &0.874  &0.083  &25.03  &0.753  &0.167  &31.75  &0.949  &0.039  &26.74  &0.769  &0.132\\
            & Ours &28.99  &0.881  &0.068  &25.40  &0.775  &0.128  &31.77  &0.951  &0.033 &27.12  &0.785  &0.102 \\
            \specialrule{0.08em}{1pt}{1pt}
            \bottomrule
        \end{tabular}
    }
    \vspace{-3em} 
    \end{center}
\end{table*}

\subsection{Experimental Settings}

\textbf{Datasets and benchmarks.} We conducted a thorough evaluation of our method across 15 real-world scenes, encompassing all scenes from the Mip-NeRF 360 dataset \cite{mipnerf360} (comprising 8 scenes) and the LLFF dataset \cite{llff} (comprising 7 scenes). These datasets are widely recognized in the field of 3D reconstruction for their diverse range of scenes, which vary in complexity and detail. The Mip-NeRF 360 dataset, in particular, provides a comprehensive platform for assessing our method's capability to capture intricate geometric and appearance details. Meanwhile, the LLFF dataset is utilized to simulate scenarios with limited viewpoints, presenting a challenging environment to test our method's proficiency in recovering scene geometry and appearance under constrained visual information.

\textbf{Evaluation metrics.} We evaluated the quality of reconstruction using three key metrics: PSNR$\uparrow$, SSIM$\uparrow$\cite{ssim}, and LPIPS$\downarrow$\cite{lpips}. PSNR is a measure of pixel-level errors, providing insight into the accuracy of the reconstruction, although it does not fully align with human visual perception as it treats all errors uniformly as noise. SSIM, on the other hand, accounts for structural transformations in luminance, contrast, and structure, offering a metric that more closely mirrors human perception of image quality. LPIPS employs a pre-trained deep neural network to extract features and assess high-level semantic differences between images, providing a similarity measure that is more aligned with human perceptual assessment compared to PSNR and SSIM.

\textbf{Implementation details.}Our method uses the 3DGS framework as a baseline, and we maintain consistency with the original 3DGS parameters, using the default settings provided by 3DGS. Specifically, the parameter \textbf{m} in Eq.\ref{eq:m} is set to 0.25, ensuring that the Geometric Attention Module is fully effective before 3DGS ceases its densification process.The Sparse-Large-Variance (SLV) initialization strategy\cite{raings} is designed to enhance point cloud reconstruction by initializing the Gaussian set with fewer components and larger variances. This approach prioritizes the learning of low-frequency components, thereby facilitating the early capture of the overall structure of the target point cloud during the optimization process. Given its alignment with our objectives, we employ SLV for random point cloud initialization in all subsequent experiments.All experiments are conducted on NVIDIA A100 GPUs, each equipped with 40GB of memory. 

\begin{figure}[hb!]
  \centering
  \includegraphics[width=0.9\linewidth]{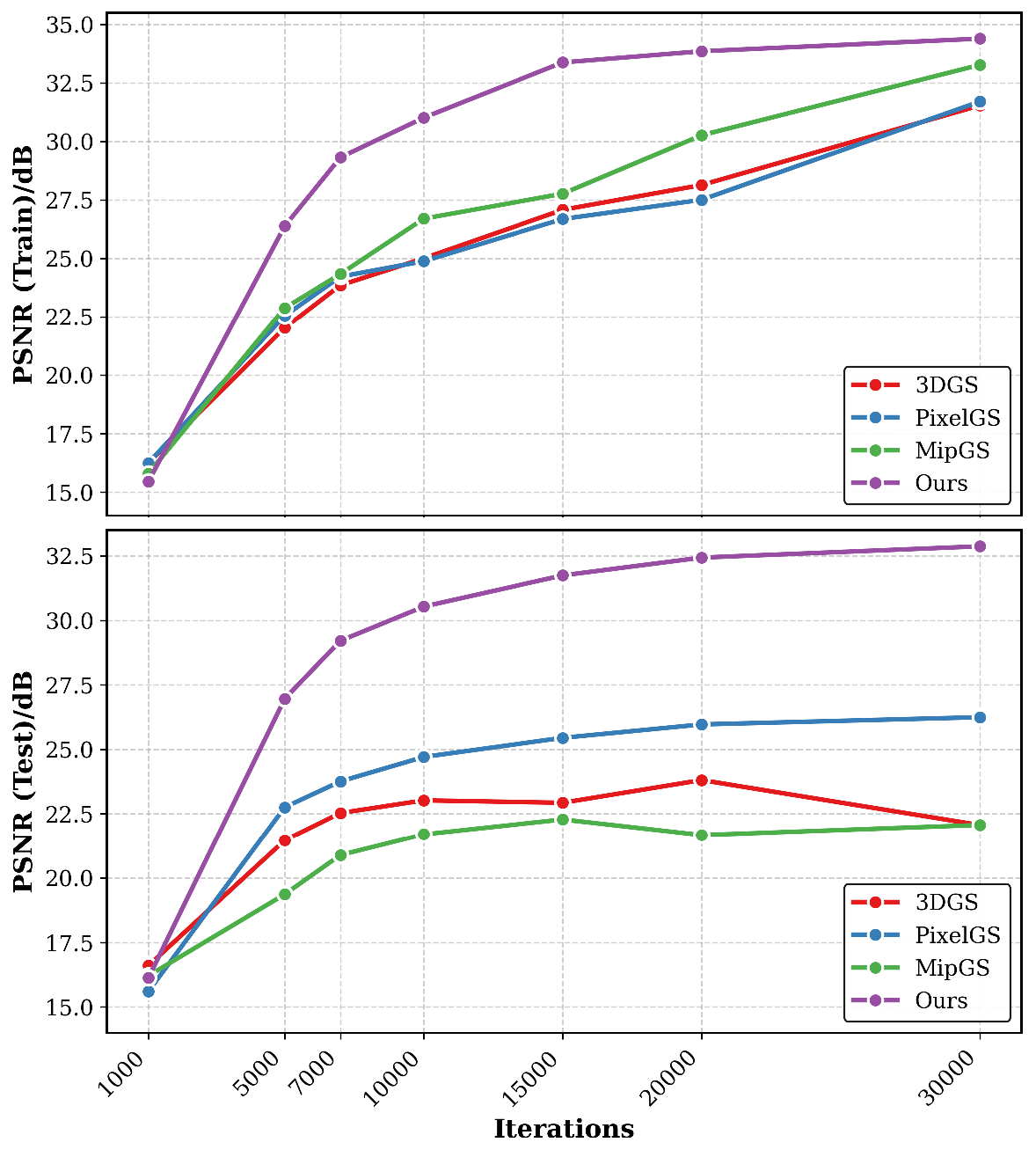}
  \caption{PSNR Evolution in Bonsai Scene Under Random Initialization.}
  \label{fig:speed}
\end{figure}

\subsection{Quantitative Results}

We compared our method against several representative approaches, including 3DGS\cite{3dgs}, Pixel-GS\cite{pixelgs}, and Mip-splatting\cite{mipgs}. All comparative methods were implemented using their official versions, and we adhered to the same training/testing split as Mip-NeRF 360, selecting one out of every eight photos for testing.

\textbf{Main results.}Table \ref{tab:main_result} presents the quantitative evaluation on the Mip-NeRF 360 dataset, highlighting the metrics of PSNR, SSIM, and LPIPS. Under random point cloud initialization, our method demonstrates a notable PSNR enhancement of 3.84, an SSIM increase of 0.088, and a LPIPS reduction of 0.084 compared to the 3DGS approach. Similarly, with noisy initialization, our method achieves a PSNR improvement of 1.70, an SSIM increase of 0.048, and a LPIPS reduction of 0.049. In three particularly challenging scenes, our approach basically surpasses other methods across all metrics, delivering performance comparable to or exceeding that of 3DGS initialized with COLMAP. Although MipGS exhibits slightly superior qualitative results in the bicycle scene, it significantly underperforms in other scenarios, underscoring the enhanced robustness of our method in handling random point cloud and noise initialization conditions. These substantial improvements highlight the robustness and efficacy of our approach in reconstructing scenes from random point clouds, ensuring reliable scene reconstruction even under challenging conditions.

\textbf{Analysis of Structural Recovery Impact.}Fig.\ref{fig:speed} showing the PSNR evolution for both training and testing sets across iterations for the Bonsai scene under random initialization conditions. Notably, methods such as 3DGS, PixelGS, and MipGS exhibit test PSNR values that stagnate prematurely, reflecting a propensity to overfit the training data. This overfitting arises because these methods, failing to recover structural information early in the optimization process, disproportionately focus on optimizing color attributes. As a result, they are unable to facilitate accurate structural recovery, leading to a disparity between training and testing performance.

In contrast, our method demonstrates a consistent trend between train and test PSNR values across iterations. This consistency underscores the efficacy of our approach in achieving early structural recovery, thereby mitigating overfitting. Ensuring robust generalization to unseen data is particularly vital in novel view synthesis tasks, where the capability to accurately predict and render new perspectives is paramount. By employing a structured attention mechanism, our method adeptly balances the optimization of both color and structural attributes, culminating in superior performance across both training and testing datasets.

\begin{table}[hb]
    \scriptsize
    \caption{Quantitative results on the Mip-NeRF 360  and LLFF dataset. For each metric, the best in {\textbf{bold}}.}
    \vspace{-2mm}
    \label{tab:mean_result}
    \begin{center}
    \setlength\tabcolsep{2pt}
    \setlength{\belowcaptionskip}{0pt}
    \renewcommand\arraystretch{1.2}
    \scriptsize
    \vspace{-1.0em}
    \resizebox{\linewidth}{!}{
        \begin{tabular}{l|ccc|ccc}
            \toprule
            & \multicolumn{3}{c}{Mip-NeRF 360} & \multicolumn{3}{c}{LLFF}\\
            & PSNR$\uparrow$ & SSIM$\uparrow$ & LPIPS$\downarrow$ & PSNR$\uparrow$ & SSIM$\uparrow$ & LPIPS$\downarrow$ \\
            \midrule
            3DGS(Random) &24.89&0.778&0.167&10.68&0.396&0.624 \\
            Ours(Random) &28.79&0.867&\textbf{0.078}&\textbf{23.18}&0.739&\textbf{0.170} \\
            3DGS(Colmap) &\textbf{28.86}&\textbf{0.874}&0.083&22.51&\textbf{0.748}&0.182 \\
            \specialrule{0.08em}{1pt}{1pt}
            \bottomrule
        \end{tabular}
    }
    \vspace{-3em} 
    \end{center}
\end{table}

\textbf{Generalization validation experiment.}Table \ref{tab:mean_result} presents a comparative analysis of 3DGS with random point cloud initialization, 3DGS with COLMAP initialization, and our method with random point cloud initialization. The results demonstrate that on both the Mip-NeRF 360 and LLFF datasets, our method significantly outperforms 3DGS with random initialization and performs comparably to 3DGS initialized with COLMAP. Notably, on the LLFF dataset, our method significantly surpasses 3DGS. This is attributed to the LLFF dataset's limited training frames and narrow viewpoints, which pose challenges for 3DGS in accurately reconstructing structures. These findings illustrate our method's capability to reconstruct structures under constrained viewpoints and its consistent generalization performance across multiple datasets.

Through these comprehensive comparisons, we validate the superior performance of our proposed method under various initialization conditions, particularly in complex and limited-view scenarios. This demonstrates its comprehensive advantages in geometric and appearance recovery, establishing it as a robust and versatile solution for real-time 3D reconstruction tasks.

\subsection{Qualitative Results.}
\begin{figure*}[h!]
  \centering
  \includegraphics[width=0.81\linewidth]{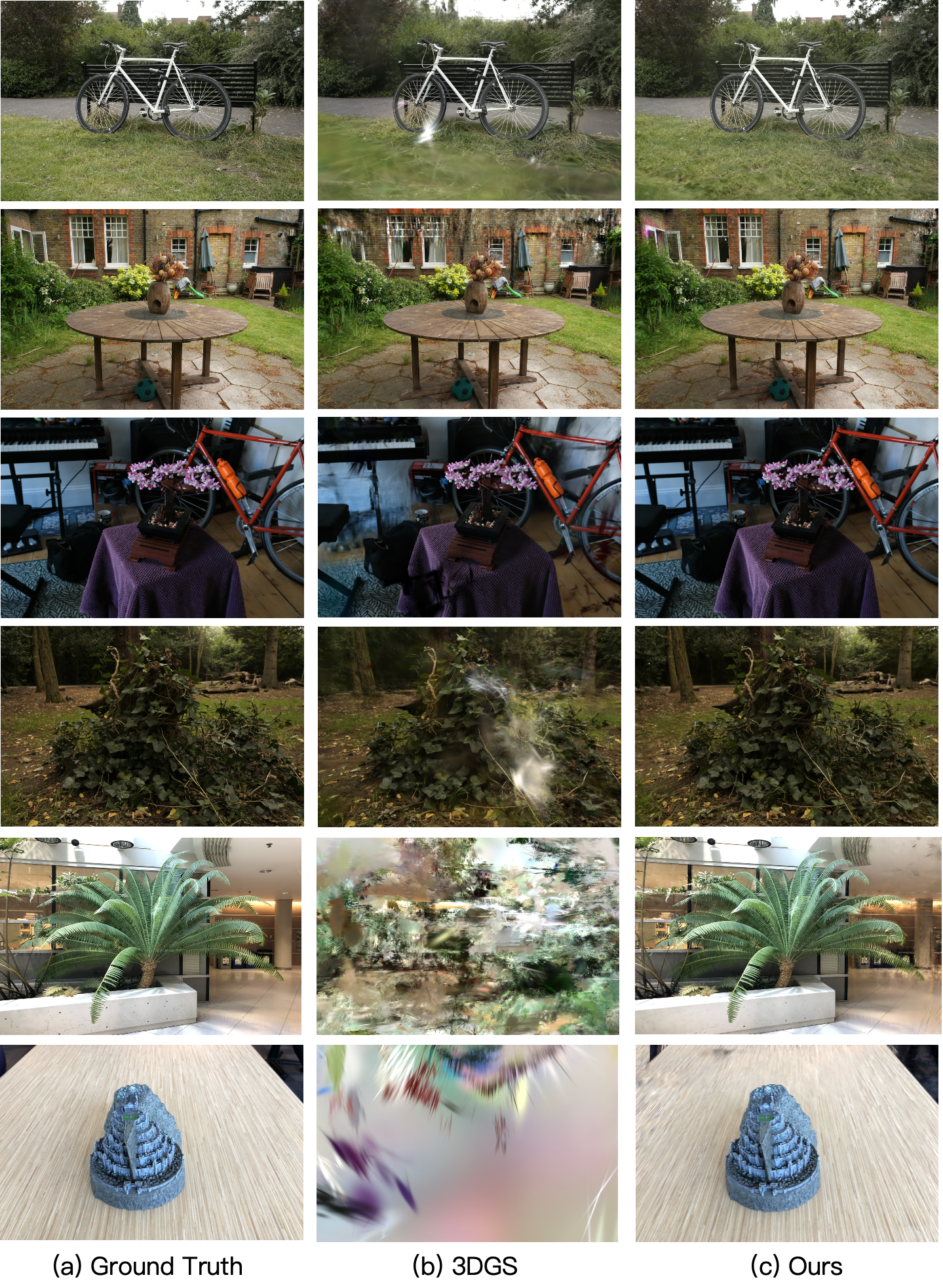}
  \caption{Qualitative comparison between Attention-GS (Ours) and 3DGS on the Mip-NeRF 360 and LLFF Dataset.}
  \label{fig:main}
\end{figure*}

In this section, we present the qualitative results from the Mip-NeRF 360 and LLFF datasets, as illustrated in Fig.\ref{fig:main}. These results compare the rendering capabilities of our proposed method against the 3DGS technique, both under random initialization conditions. Our approach consistently outperforms in recovering intricate details and complex structures, underscoring its robustness and accuracy.

For the Mip-NeRF 360 dataset, in the Bicycle scene, our method excels in precisely reconstructing small and intricate structures, such as the fine details of grass. This highlights our method's proficiency in capturing subtle textures and elements that are often challenging to render accurately. In the Garden scene, our approach effectively restores the planar structures of distant houses, demonstrating its ability to maintain structural integrity and spatial coherence even in complex environments. This capability is particularly significant as it showcases our method's strength in handling large-scale scenes with varying depth and perspective. Furthermore, in the Bonsai scene, our method surpasses 3DGS in reconstructing geometric structures, such as the legs of a piano. 

For the LLFF dataset, we focused on the Fern and Fortress scenes, which are characterized by limited frames and narrow viewpoints. These conditions pose significant challenges for 3DGS, which struggles to reconstruct the scene structures accurately from random point clouds. In contrast, our method effectively overcomes these limitations, successfully reconstructing the intricate details of the Fern scene's foliage and the architectural integrity of the Fortress scene. This demonstrates our method's robustness in handling scenarios with sparse data and constrained perspectives, ensuring high-fidelity novel view rendering even under challenging conditions.

These qualitative findings strongly align with our quantitative analysis, further corroborating the efficacy of our method. 

\begin{table*}[h!]
    \scriptsize
    \caption{Quantitative results on the Mip-NeRF 360 dataset. For each metric, the best in {\textbf{bold}}.}
    \vspace{-2mm}
    \label{tab:ab_study}
    \begin{center}
    \setlength\tabcolsep{2pt}
    \setlength{\belowcaptionskip}{0pt}
    \renewcommand\arraystretch{1.2}
    \scriptsize
    \vspace{-1.0em}
    \resizebox{\linewidth}{!}{
        \begin{tabular}{l|ccc|ccc|ccc|ccc}
            \toprule
            & \multicolumn{3}{c}{Mip-NeRF 360(Avg.)} & \multicolumn{3}{c}{Bycicle} & \multicolumn{3}{c}{Bonsai} & \multicolumn{3}{c}{Stump}\\
            & PSNR$\uparrow$ & SSIM$\uparrow$ & LPIPS$\downarrow$ & PSNR$\uparrow$ & SSIM$\uparrow$ & LPIPS$\downarrow$ & PSNR$\uparrow$ & SSIM$\uparrow$ & LPIPS$\downarrow$ & PSNR$\uparrow$ & SSIM$\uparrow$ & LPIPS$\downarrow$\\
            \midrule
            3DGS &24.89  &0.778  &0.167  & 22.06 & 0.562 & 0.366 &23.96  &0.841  &0.126 & 21.19 &0.609  &0.249\\
            3DGS+Geo &26.67  &0.791  &0.088  &24.92  &0.723 &0.203  &31.43  &0.945  &0.041 &25.11  &0.722  &0.121 \\
            3DGS+Geo+Opacity-grad &28.08  &0.855  &0.087  & 25.08 & 0.732 &0.185 &31.64  &0.946  &0.038 & 26.29 &0.759  &0.109\\
            Ours &\textbf{28.73}  &\textbf{0.866}  &\textbf{0.083}  &\textbf{25.13}  &\textbf{0.743}  &\textbf{0.167}  &\textbf{32.82}  &\textbf{0.953}  &\textbf{0.031} &\textbf{26.47}  &\textbf{0.761}  &\textbf{0.108}\\
            \specialrule{0.08em}{1pt}{1pt}
            \bottomrule
        \end{tabular}
    }
    \vspace{-3em} 
    \end{center}
\end{table*}
\subsection{Ablation Studies}

To assess the effectiveness of each component in our method, we conducted ablation studies on the Mip-NeRF 360 dataset, with results detailed in Table \ref{tab:ab_study}. By incrementally integrating each component, we evaluated their individual contributions to overall performance, especially under random point cloud initialization conditions.

The results reveal a consistent improvement in evaluation metrics as components are added, underscoring the independent effectiveness of each module in enhancing model performance and recovering geometric structures from randomly initialized point clouds. Starting with the baseline 3DGS, we recorded a PSNR of 24.89, SSIM of 0.778, and LPIPS of 0.167. Introducing the Geometric Attention module increased PSNR to 26.67, SSIM to 0.791, and reduced LPIPS to 0.088, highlighting its capability in capturing spatial relationships and aiding geometric structure recovery. Further enhancement was observed with the addition of Opacity-weighted Gradients, boosting PSNR to 28.08, SSIM to 0.855, and reducing LPIPS to 0.087. Our complete method achieved the highest performance, with a PSNR of 28.73, SSIM of 0.866, and LPIPS of 0.083, confirming the synergistic effect of all components in enhancing model robustness and accuracy. In the Bonsai scene, our method achieved a PSNR of 32.82 and an SSIM of 0.953, significantly outperforming the baseline, demonstrating superior capability in handling complex geometries and diverse appearances from random point clouds.

The ablation study validates the independent and combined advantages of each component, proving the method's applicability and robustness in complex 3D scene reconstruction tasks. These experiments highlight our method's comprehensive performance benefits, particularly in recovering accurate geometric structures and appearances from random initializations.

\section{Conclusion}
In this paper, we presented AttentionGS, a robust framework that addresses the limitations of traditional 3D Gaussian Splatting methods, particularly their dependency on high-quality initial point clouds. By integrating a geometry attention-guided global structure reconstruction method, an appearance attention-driven adaptive optimization strategy, and an opacity-aware densification criterion, our approach significantly enhances rendering quality and structural fidelity. Extensive experiments on standard datasets demonstrate that AttentionGS consistently outperforms existing methods, especially under challenging conditions such as random initialization of point clouds. Our method not only excels in recovering details and structures with poor-quality point clouds, proving its robustness and accuracy, but also extends the applicability of 3DGS in the domain of novel view synthesis by eliminating the reliance on initial point cloud quality. This advancement broadens the scope of 3DGS applications, making a meaningful contribution to the field of 3D vision.

\bibliographystyle{ACM-Reference-Format}
\bibliography{main.bib}
\end{document}